\documentclass[a4paper]{article}
\usepackage{INTERSPEECH2019}
\usepackage[utf8]{inputenc} % allow utf-8 input
\usepackage[tracking=true]{microtype}
\usepackage{subfig}
\usepackage{wrapfig}

\usepackage{graphicx}
\usepackage{amssymb,amsmath,bm, caption}
\usepackage{textcomp}
\usepackage{mathtools}
\def\vec#1{\ensuremath{\bm{{#1}}}}

\usepackage{multirow}
\DeclareMathOperator\LayerNorm{LayerNorm}
\DeclareMathOperator\SelfAttention{SelfAttention}
\DeclareMathOperator\Activation{Activation}
\usepackage{color}

\usepackage{hyperref}
\usepackage{cite}  % cite numeric ranges
\usepackage{enumitem}
\setlist{
	itemsep=0pt,
	parsep=0pt plus 2pt minus 1pt,
	topsep=0pt plus 2pt minus 1pt,
	partopsep=0pt}

\makeatletter

\renewcommand{\section}{\@startsection
  {section}%
  {1}%
  {}%
  {-0.3\baselineskip}%
  {0.3\baselineskip}%
  {}}%

\renewcommand{\subsection}{\@startsection
  {subsection}%
  {2}%
  {}%
  {-0.1\baselineskip}%
  {0.1\baselineskip}%
  {}}%

\renewcommand{\subsubsection}{\@startsection
  {subsubsection}%
  {3}%
  {}%
  {-0.1\baselineskip}%
  {0.1\baselineskip}%
  {}}%

\g@addto@macro\normalsize{%
  \setlength\abovedisplayskip{3pt plus 2pt minus 2pt}
  \setlength\belowdisplayskip{3pt plus 2pt minus 2pt}
  \setlength\abovedisplayshortskip{3pt plus 2pt minus 2pt}
  \setlength\belowdisplayshortskip{3pt plus 2pt minus 2pt}
}

\makeatother

\ninept
\title{Language Modeling with Deep Transformers}
\name{Kazuki Irie$^1$, Albert Zeyer$^{1,2}$, Ralf Schl\"uter$^{1}$, Hermann Ney$^{1,2}$}
\address{$^1$Human Language Technology and Pattern Recognition Group, Computer Science Department \\
RWTH Aachen University, 52074 Aachen, Germany\\
$^2$AppTek GmbH, 52062 Aachen, Germany}
\email{\{irie, zeyer, schlueter, ney\}@cs.rwth-aachen.de}

\normalsize

\begin{document}
  \maketitle
\begin{abstract}
\vspace{2mm}
We explore deep autoregressive Transformer models in language modeling for speech recognition.
%Our contribution is two folds:
%Our work focuses on two aspects.
%Our work presents two special focuses.
We focus on two aspects.
First, we revisit Transformer model configurations specifically for language modeling.
We show that well configured Transformer models outperform our baseline models based on the
shallow stack of LSTM recurrent neural network layers.
We carry out experiments on the open-source LibriSpeech 960hr task, for both 200K vocabulary word-level
and 10K byte-pair encoding subword-level language modeling.
We apply our word-level models to conventional hybrid speech recognition by lattice rescoring, and the
subword-level models to attention based encoder-decoder models by shallow fusion.
%where we achieve slightly better results than previously reported best performance for end-to-end models.
%We confirm results on the Switchboard 300hr task.
Second, we show that deep Transformer language models do not require positional encoding. % because of autoregression.
The positional encoding is an essential augmentation for the self-attention mechanism which is
invariant to sequence ordering.
However, in autoregressive setup, as is the case for language modeling, the amount of information increases along the position dimension, which is a positional signal by its own.
%The analysis of attention weights shows that when the number of layers is large enough, autoregressive self-attention models can
The analysis of attention weights shows that deep autoregressive self-attention models can
automatically make use of such positional information.
We find that removing the positional encoding even slightly improves the performance of these models.
%We report perplexities and WERs for both word level conventional speech recognition and
%subword level end-to-end speech models with state-of-the-art performance. 
\end{abstract}
\vspace{-1mm}
\section{Introduction}
%\vspace{-1mm}
%The \textit{Transformer} encoder-decoder model \cite{transfo} has been originally proposed for
%machine translation and constituency parsing.
Transformer encoder-decoder models \cite{transfo} have become popular in natural language processing.
The Transformer architecture allows to successfully train
a deep stack of \textit{self-attention} layers \cite{cheng16,lin2017structured,ParikhT0U16}
via residual connections \cite{resnet} and layer normalization \cite{ba2016layer}.  % cite papers beyond NLP
The positional encodings \cite{transfo, GehringAGYD17}, typically based on sinusoidal functions,
are used to provide the self-attention with the sequence order information.
%Across various applications, even beyond NLP, systematic improvements have been reported
Across various applications, systematic improvements have been reported
over the standard, multi-layer long short-term memory (LSTM) \cite{hochreiter1997long}
recurrent neural network based models. % which used to be the dominant approach for sequence modeling.
%in natural language processing and beyond.
%Recent success of self-attention based model "Transformer" \cite{transfo} in various natural language processing tasks
While originally designed as an encoder-decoder architecture in machine translation,
the \textit{encoder} (e.g., \cite{devlin2018bert}) and the \textit{decoder} (e.g., \cite{liu2018generating}) components
are also separately used in corresponding problems
depending on whether the problem disposes the whole sequence for prediction or not.
%depending on whether the problem is autoregressive or the whole sequence is available for prediction.

A number of recent works have also shown impressive performance in language modeling
using the Transformer \textit{decoder} component
\cite{liu2018generating, dai2019, al2018character, baevski2018adaptive, radford2018improving, gpt2}.
The earliest example can be found in \cite{liu2018generating}
where such models are investigated for text generation.
Recent works on training larger and deeper models \cite{al2018character, radford2018improving, gpt2} have
shown further potential of the Transformer in language modeling. On the other hand, an obvious limitation of the
Transformers is that their memory requirement linearly increases in terms of number of tokens in the sequence, which
requires to work with a limited context window
(basically a $n$-gram model where the typical number for $n$ is 512) for tasks dealing
with long sequences such as character-level language modeling \cite{al2018character}.
Dai et al.~\cite{dai2019} has introduced a segment-level recurrence and relative positional encoding in the Transformer
language model to be able to potentially handle unlimited context.

In this work, we investigate deep autoregressive Transformers for language modeling in speech recognition.
To be specific, we focus on two aspects.
First, we revisit the parameter configurations of Transformers,
originally engineered for the sequence-to-sequence problem \cite{transfo},
specifically for language modeling.
We conduct experiments on the LibriSpeech automatic speech recognition (ASR) task \cite{panayotov2015librispeech}
for both word-level conventional speech recognition and byte-pair encoding (BPE) \cite{sennrich16bpe} level end-to-end speech recognition \cite{lasicassp2016, zeyer2018:asr-attention}.
We apply our word-level models to hybrid speech recognition by \textit{lattice rescoring} \cite{sundermeyer2014:rescoring},
and the BPE-level models to end-to-end models by \emph{shallow fusion} \cite{gulcehre+al-2016-monolingual, toshniwal2018comparison}.
We show that well configured Transformer language models
outperform models based on the simple stack of LSTM RNN layers in terms of both
perplexity and word error rate (WER).
%\TODO sota wers. We also confirm results on the Switchboard 300hr results.

%Second, we closely look at properties of the autoregressive self-attention and experimentally show that the standard positional encoding is
%not needed for language modeling.
Second, we experimentally show that the positional encoding is not needed for multi-layer autoregressive self-attention models.
The visualization of the attention weights shows that when the sinusoidal positional encoding is provided with the input,
the first layer of the Transformers learns to extract $n$-gram features (therefore making use of positional information).
However, in the autoregressive problem where a new token is provided to the model at each time step, the amount of information the model has access to strictly increases from left to right at the lowest level of the network, which should provide some positional information by its own.
We observe that deep Transformer language models without positional encoding automatically make use of such information, and even
give slight improvements over models with positional encodings.% (typically by 4\% relative perplexity).
%We also note that for these models to make use of this autoregressive property
%of the problem, it requires to be somewhat deep. % (in our experiments, 12 works, 6 does not).
%We observe that it requires a few layers to build position-aware features and again a few layers above to make use of such features.
%(\TODO speculation or is this clear from the plots?)
%LibriSpeech \cite{panayotov2015librispeech}
%\TODO trace the exact literatures, clarify paths from self-attention to deep transformer. first self-attention in \cite{cheng16} (already for %language modeling but shallow) in
%fact also related to \cite{tran2016recurrent, irie16}. Then used by Google \cite{ParikhT0U16} and IBM \cite{lin2017structured}
%\label{sec:intro}
%\vspace{-1mm}
\vspace{-0.5mm}
\section{Related Work}
\vspace{-0.5mm}
%Many recent works investigate the potential of Transformer decoder architecture for language modeling \cite{liu2018generating, dai2019, al2018character, radford2018improving, gpt2}.
The first part of our work follows the spirits of Al-Rfou et al.'s work \cite{al2018character} and
Radford et al.'s work \cite{radford2018improving, gpt2} in investigating larger and deeper Transformers
for language modeling. We show that deep Transformer language models can be successfully applied to speech recognition and give good performance.
The second part of this work concerns the positional encoding, which is a crucial component in the original Transformer.
A number of previous work investigated positional encoding variants to improve self-attention (e.g., \cite{ShawUV18, SperberNNSW18, salazar19, dai2019}).
%There are active investigations on positional encoding variants to improve self-attention (e.g., \cite{ShawUV18, SperberNNSW18, salazar19, dai2019}).
Previous works in Transformer language models systematically use positional encoding, either jointly learned one or the sinusoidal one
(both cases are reported to give similar performance in \cite{al2018character}).
We show that the deep autoregressive self-attention models do not require any explicit model for encoding positions to give the best performance.
\vspace{-0.5mm}
\section{Autoregressive Self-Attention}
%\vspace{-0.5mm}
\label{sec:model}
The language model we consider is based on the \textit{decoder component} of the Transformer architecture \cite{transfo}.
Similar to previous work \cite{liu2018generating, radford2018improving, baevski2018adaptive, al2018character, dai2019, gpt2},
we define \textit{layer} as a stack of two components: \textit{self-attention} and \textit{feed-forward}\footnote{Typically called
\textit{position-wise feed-forward} module \cite{transfo}. Here we omit \textit{position-wise} as it is obvious for autoregressive models.}
modules.

The autoregressive \textit{self-attention module} in the $l$-th layer transforms the input $z_t^{(l-1)}$ at position $t$ as follows:
   \begin{eqnarray*}
   \notag
x_t^{(l)} &=& \LayerNorm(z_t^{(l-1)}) \\ \notag
q_t^{(l)}, k_t^{(l)}, v_t^{(l)} &=& Q x_t^{(l)}, K x_t^{(l)}, V x_t^{(l)} \\
h_t^{(l)} &=& \big(h_{t-1}^{(l)}, (k_t^{(l)}, v_t^{(l)})\big) \\ \notag
y_t^{(l)} &=& z_t^{(l-1)} + W_{0} \SelfAttention(h_t^{(l)}, q_t^{(l)})
    \end{eqnarray*}
where $Q$, $K$, $V$, respectively denote query, key, value projection matrices, $\LayerNorm$ denotes layer normalization \cite{ba2016layer},
$\SelfAttention$ denotes the scaled multi-head dot product self-attention \cite{transfo}, and $W_0$ denotes the projection matrix for
the residual connection \cite{resnet}.
%In our experiments, we use the same dimension for the value projection as the key and query dimension as well as for the residual connection.

The output $y_t^{(l)}$ is then fed to the \textit{feed-forward} module:
   \begin{eqnarray*}
m_t^{(l)} &=& \LayerNorm(y_t^{(l)}) \\
z_t^{(l)} &=& y_t^{(l)} + W_{2}\Activation(W_{1} m_t^{(l)})
    \end{eqnarray*}
where $\Activation$ is rectifier \cite{nair10}, Gaussian error linear unit (GELU) \cite{gelu, gpt2}, or gated linear unit (GLU) \cite{dauphin2017lm} in this work.
The final model is build by stacking these \textit{layers} multiple times.
%Since we use the same dimension
%specifications across all layers, we can specify our models by the tuple (number of layers $L$, feed-forward dimension $d_{ff}$, residual dimension $d_{res}$, number of heads $H$).

The input of the network consists of the sum of the token embedding (word or BPE in this work) and the sinusoidal \textit{positional
encoding} as specified in \cite{transfo}. The output softmax layer gives
the probability distribution for the next token.
As shown in the equations above, $h_t^{(l)}$ can be seen as \textit{states} of the Transformer model\footnote{In principle, we
could also consider an autoregressive self-attention model which updates states at all predecessor positions for each new input, which would be then much more computationally inefficient.}
(whose size, as opposed to the RNN states, linearly grows along the position dimension). During inference,
these states are stored to avoid redundant computation.
During training, the computation along the position dimension is parallelized for speed-up.

\section{LibriSpeech Dataset}
\subsection{Language Modeling Data Descriptions}
The LibriSpeech datasets \cite{panayotov2015librispeech} for language modeling consists of 800M-word text only data
and 960hr of audio transcriptions which corresponds to 10M-word text data.
Based on analysis of count model perplexities, we observe that the audio transcription
part does not contain special domain signal which matches the development set. Therefore,
we simply merge the two datasets to form a single dataset for language model
training.
The average sentence length in the resulting training data is 21 words with the maximum length of 600 words.
The development and test sets respectively have two parts\cite{panayotov2015librispeech}: dev-clean, dev-other, test-clean, and test-other.
This separation is based on the audio-level characteristics,
therefore it has no special meaning for language modeling.
In the experimental section, we denote by "\textbf{Dev}" and "\textbf{Test}"
the concatenation of \textit{clean} and \textit{other} parts of the respective data.
Both datasets consist of about 110K running words with average of 20 words per sentence.
The word-level vocabulary contains 200K words.
We report all perplexities without making use of contexts beyond the sentence boundary.

\subsection{4-gram count and LSTM-RNN Baselines}
\label{sec:base}
We use the official 4-gram count language model provided with the LibriSpeech dataset \cite{panayotov2015librispeech}.
No improvement in perplexity is observed when going up to 5-grams.
For LSTM-RNN language models \cite{sundermeyer2012lstm}, we first train our base configuration;
the model has 2 LSTM-RNN layers with 2048 nodes and the input projection layer of 128, where the dropout
with a rate of 0.2 is applied between each layer.
Since we observe that this model underfits the LibriSpeech training set, we remove the dropout and
further increase the model size, which effectively give better perplexities as shown in Table \ref{tab:base}.
We find that improvements from simply stacking layers saturate at 4 layers even without overfitting. Introducing a small linear bottleneck
layer (size 512 here) before the output layer can make the models compact but with a loss in performance.
The best model we obtain has 2 layers with 4096 nodes.
Relative improvements greater than 58\% are obtained by the LSTM over the 4-gram language model.

\begin{table}[!ht]
	\vspace{-1mm}
	\caption{\label{tab:base} {\it Perplexities of the \textbf{baseline} models.}}
	\vspace{-1mm}
	\begin{center}
		\setlength{\tabcolsep}{0.3em}
		%      \resizebox{\columnwidth}{!}{
		\begin{tabular}{|l|c|c||c|c|c||r|r|} \hline
			\multirow{2}{*}{Model} & Drop- & Bottle-& Num.  & Num. & Params & \multirow{2}{*}{Dev} & \multirow{2}{*}{Test} \\
			                       & out   & beck   & units & layers & in M   &   &   \\ \hline
			4-gram    & - & - & - & - & 230 & 146.2 & 151.8 \\ \hline
			\multirow{10}{*}{LSTM}     & 0.2  & \multirow{8}{*}{None} & \multirow{7}{*}{2048} & \multirow{2}{*}{2} & \multirow{2}{*}{487} & 71.3 & 74.8  \\ \cline{2-2} \cline{7-8}
			& \multirow{9}{*}{0.0}   & & & &  & 66.6 & 69.9  \\ \cline{5-5} \cline{6-6}
			&  & & &3 & 520  & 64.0 & 67.2 \\ \cline{5-5} \cline{6-6}
			&  & & &4 & 554  & \textbf{61.9} & \textbf{64.9} \\ \cline{5-5}  \cline{6-6}
			&  & & &5 & 587  & 62.7 & 65.9 \\ \cline{5-5}  \cline{6-6}
			&  & & &6 & 621  & 64.5 & 67.5  \\ \cline{5-5}  \cline{6-6}
			&  & & &8 & 688  & 67.2 & 70.3 \\ \cline{4-5} \cline{6-8}
			& &  & \multirow{2}{*}{4096} & \multirow{2}{*}{2} & 1048 & \textbf{60.2} & \textbf{63.2} \\ \cline{3-3} \cline{6-8} %\cline{2-4} \cline{6-8}
			% BOUNDARY?
			& & \multirow{2}{*}{512} &   &  & 334  & 63.1 & 66.3 \\ \cline{4-6}
			&  &  & 2048 & 4 & 248 & 64.5 & 67.7 \\ \hline
			%                                          & 4 & 2048 & \multirow{2}{*}{512} & & 248 & 64.5 & 67.7 \\ \cline{3-3} \cline{6-8}
			%                                          & 2 & 4096 &  &  & 334  & 63.1 & 66.3 \\ \hline
		\end{tabular}
		%      }
	\end{center}
	\vspace{-3mm}
\end{table}
\section{Text based Experiments}
We carry out experiments for both word-level and BPE-level language modeling.
We first focus on the word-level one.
%\vspace{-2mm}
%\TODO perplexity break down for clean and other parts.
\subsection{Hyper-parameters in Transformers}
The Transformer architecture is a new search space Odyssey \cite{GreffSKSS17}.
The exhaustive model hyper-parameters for Transformer language models specified by the
equations in Sec.~\ref{sec:model} are the input token embedding size, the number of layers, the dimension of the residual connection, and for each layer the number of attention heads, the dimension of the key and query, the dimension of the value, and the dimension of the feed-forward layer.

In our experiments, we use the same dimension for key, query and value, as well as the residual connection. We
use the same dimensionality across all layers. Therefore, our models can be fully specified by
the tuple (\textbf{number of layers $L$, feed-forward dimension $d_{ff}$, residual dimension $d_{res}$, number of heads $H$}).
We do not apply any regularization method including dropout.
We train all models using the plain stochastic gradient descent and new-bob learning rate tuning on a single GPU.
We define our training sub-epoch (for new-bob) as the 10th of the full training data.
All our implementations are based on the Tensorflow \cite{tfshort} based open-source toolkit RETURNN \cite{zeyer2018:returnn}\footnote{Training configuration files and trained models are available at \scriptsize \url{https://github.com/rwth-i6/returnn-experiments/tree/master/2019-lm-transformers}.}.

\subsection{Hyper-parameter Tuning}
Given the amount of LibriSpeech training data (810M words), it is unreasonable to train all model variants until full convergence.
The earlier stage of the training already consistently indicates the potential performance of the models.
Therefore, we first carry out comparisons between models with different configuration at the equal, large enough, but reasonable number of updates.

The first set of comparison investigates the effect of depth and width. The perplexity results can be found in Table \ref{tab:depth_ep25}. All models in the table use 8 attention heads. Other parameters are specified in the table.
The table is organized in three parts.
The upper part of Table \ref{tab:depth_ep25} shows the effect of number of layers; we observe
that increasing number of layers (therefore the number of parameters) from 1 to 42 gradually improves the perplexity.
In the middle part of Table \ref{tab:depth_ep25} , we vary both the number of layers, feed-forward dimension, and the residual dimension.
First of all, the 12-layer $(12, 4096, 512, 8)$ model outperforms the 6-layer $(6, 8192, 512, 8)$ model, while having similar number of parameters, which seems to indicate that the depth effectively benefits Transformer language models.
We also train an extreme model which has only 2 layers with wide dimensions $(2, 8192, 2048, 8)$. The number of parameters in fact blows up because of the large value of $d_{res}$  which results in a large
matrix in the output softmax layer with 200K vocabulary\footnote{
We note that this is also the reason why the number of parameters of our baseline LSTM language models in Table \ref{tab:base} is relatively high.}.
We observe that such wide but shallow models do not perform well\footnote{Since the softmax bottleneck dimension typically needs to be large for the best performance (\cite{yang2017breaking}; Table \ref{tab:base}), we also train a (12, 2048, 512, 8) model
	where we insert an additional projection layer with a large dimension (2048)
	before the output layer; no improvement was obtained.
	}.
Finally, the lower part of Table \ref{tab:depth_ep25} shows deeper models with a smaller input dimension.
	\vspace{-2mm}
\begin{table}[!ht]
	\vspace{-1mm}
	\setlength{\tabcolsep}{0.3em}
	\caption{\label{tab:depth_ep25} {\it Perplexity \textbf{after 2.5 epoch} (25 sub-epochs in our setup; 6.5M updates).
			The number of heads $H$} is 8 for all models below.} % 261471 updates per sub-epoch.
		\vspace{-1mm}
	\begin{center}
		%\resizebox{\columnwidth}{!}{
		\begin{tabular}{|c|c|c|c|c||r|r|} \hline
			Input&  \multirow{2}{*}{$L$} &  \multirow{2}{*}{$d_{ff}$} & \multirow{2}{*}{$d_{res}$} & Params. & \multicolumn{2}{c|}{Perplexity} \\ \cline{6-7}
			emb.& &       &     & in M & Train & \multicolumn{1}{|c|}{Dev}  \\ \hline \hline
		    \multirow{6}{*}{512}& 1   & \multirow{6}{*}{2048}  & \multirow{6}{*}{512} & 208 & 108.3 & 104.9 \\
			& 6   & & & 224 & 75.7  &  74.3    \\
			& 12  & & & 243 & 67.6  & 67.1    \\
			& 24  & & & 281 & 62.2  & 62.3   \\
			& 32	& & & 306 & 60.1  & 60.6   \\
			& 42  & & & 338 & \textbf{59.0}     &  \textbf{59.6}  \\ \hline \hline
%			& 64  & & & 407 & 57.0     & 58.2 \\  \hline \hline
%            56  & & & 382 & 55.5     &  57.3  \\ \hline \hline % check the learning rate!
			\multirow{5}{*}{512} & 2	&  \multirow{2}{*}{8192} & 2048  & 536 & 73.1 & 73.8  \\ \cline{4-4}
			& 6   &         & \multirow{4}{*}{512} & 262 & 66.7 & 66.7  \\ \cline{3-3}
			& 12  &  4096   &   & 268 & \textbf{63.5}  & \textbf{63.8}  \\ \cline{2-3}
			&\multirow{2}{*}{4}   & 16384  &  & 277 & 67.6 & 67.4 \\ \cline{3-3}
            &                     & 32768  &  & 344 & 65.4 & 68.4 \\ \hline \hline 
            \multirow{4}{*}{128}  & 64   &  \multirow{4}{*}{2048}       & \multirow{4}{*}{512} & 330 & 56.3  & 57.6   \\ 
                                  & 80   &        &  & 380 & 53.1  & 55.5  \\
                                  & 96   &                              &                      & 431 & 51.9 & 54.9 \\ 
                                  & 112  &                              &                      & 481 & \textbf{51.5} & \textbf{54.5} \\  \hline
		\end{tabular}
	\end{center}
	\vspace{-3mm}
\end{table}

%However, the softmax bottleneck dimension for language modeling typically needs to be large for the best performance \cite{yang2017breaking}. In Transformers, the bottleneck dimension corresponds to the residual connection
%dimension which is typically kept rather small (typically 512 or 1024).
%As a control experiment, we also train a model in which we insert an additional projection layer with a large dimension
%before the softmax layer to give larger bottleneck capacity.
%Table \ref{tab:bottleneck} shows the comparison conducted on the (12, 2048, 512, 8) model.
%We observe that simply enlarging the bottleneck dimension does not improve Transformer models.
%\begin{table}[ht]
%	\vspace{-4mm}
%	\caption{\label{tab:bottleneck} {\it Effect of larger softmax bottleneck dimension. Perplexity \textbf{after 1 epoch} (10 sub-epochs in our setup; 2.6M updates) for (12, 2048, 512, 8).}}
%		\vspace{-1mm}
%	%	\vspace{-7mm}
%	\begin{center}
%		%\resizebox{\columnwidth}{!}{
%		\begin{tabular}{|l|c||r|r|} \hline
%			Bottleneck  & Params.  &  \multicolumn{2}{c|}{Perplexity} \\ \cline{3-4}
%			dimension   & in M    & Train & \multicolumn{1}{|c|}{Dev}  \\ \hline
%			512  & 243 & 76.4 & 72.5  \\
%			2048 & 512 &  74.7 & 73.8  \\ \hline
%		\end{tabular}
%	\end{center}
%	\vspace{-3mm}
%\end{table}

Table \ref{tab:heads} shows the effect of number of attention heads.
16 heads which is the largest number we try in this setup give the best performance. 
In addition, we examine the type of activation function (Table \ref{tab:act}).
As opposed to previous work on feed-forward language models using GLUs \cite{dauphin2017lm, irie18}, we do not observe
faster convergence. As we observe that the impact of choice of activation functions on the perplexity
is overall limited, all our other models use the standard ReLU.
As reported in the original Transformer, we confirm that both layer normalization and residual connections are needed for these models for stable training\footnote{We tried to train multiple models without either residual connections
or layer normalization. Also, following \cite{gpt2}, we tried reorganizing the \textit{feed-forward module} to insert one additional pre-activation layer normalization \cite{HeZRS16} and one more activation function.
However, we did not observe any improvement. The original Transformers anyway do not have any activation on the residual path
throughout the whole network.}.

Finally, we train models with the best configurations for longer. Table \ref{tab:final} shows the perplexities which are better than those obtained by our LSTM based models (Table \ref{tab:base}).

    \begin{table}[!ht]
    	    	        \vspace{-2mm}
    	        	\begin{minipage}[b]{0.46\linewidth}
    	        		\caption{\label{tab:heads} {\it Effect of number of heads.~Perplexity \textbf{after 2.5 epoch} for (12, 2048, 512, $H$).}}
    	        		%	\vspace{-7mm}
    	        		\begin{center}
    	        			\resizebox{\columnwidth}{!}{
    	        				\begin{tabular}{|c|c||r|r|} \hline
    	        					\multirow{2}{*}{$H$} & Params. & \multicolumn{2}{c|}{Perplexity}\\ \cline{3-4}
    	        					&     in M         &  Train& \multicolumn{1}{|c|}{Dev}  \\ \hline
    	        					1	& \multirow{4}{*}{243}   & 71.9 & 70.8   \\ 
    	        					4   &  & 69.1  & 68.6   \\
    	        					8   &  & 67.6  &  67.1   \\ 
    	        					16  &  & \textbf{66.9}  & \textbf{66.6} \\ \hline
    	        					%			 8    & 512 & 62.2 & 62.3 & 281 \\ \hline % 24 layers
    	        					%			12   & 768 & 59.5 & 60.3 & 440  \\ \hline % 24 layers
    	        				\end{tabular}
    	        			}
    	        		\end{center}
    	        		
    	        	\end{minipage}
    \hspace{0.3cm}
    	\begin{minipage}[b]{0.46\linewidth}
    		\caption{\label{tab:act} {\it Effect of activation functions. Perplexity \textbf{after 1 epoch} (10 sub-epochs in our setup) for (24, 2048, 512, 8).}}
    		%        \vspace{-7mm}
    		\begin{center}
    			\resizebox{\columnwidth}{!}{
    				\begin{tabular}{|l|r|r|} \hline
    					\multirow{2}{*}{$\Activation$} &  \multicolumn{2}{c|}{Perplexity} \\ \cline{2-3}
    					& Train & \multicolumn{1}{|c|}{Dev} \\ \hline
    					ReLU \cite{nair10, transfo}   &  76.4 & 72.5 \\
    					GLU  \cite{dauphin2017lm}   &  76.5 & 72.8 \\
    					GELU \cite{gelu, gpt2}   &  \textbf{75.7} & \textbf{72.2} \\ \hline
    				\end{tabular}
    			}
    		\end{center}
    	\end{minipage}
    	        \vspace{-2mm}
    \end{table}
\begin{table}[!ht]
		\setlength{\tabcolsep}{0.35em}
%\vspace{-2mm}
\caption{\label{tab:final} {\it Perplexities \textbf{after longer training}.}}
%\caption{\label{tab:final} {\it Perplexities of the best models \textbf{after convergence}.}}
%\caption{\label{tab:final} {\it Perplexities after convergence of the best models. $d_{res}$ is 512 and $H$ is 8 for all models.}}
%	\caption{\label{tab:final} {\it Final Perplexities of the best models. $d_{res}$ is 512 and $H$ is 8 for all models.}}
\vspace{-1mm}
	\begin{center}
%		\resizebox{\columnwidth}{!}{
       \begin{tabular}{|c|c||c|c|c|c||r|r|r|} \hline
           Max. & Conv- & \multirow{2}{*}{$L$}  &  \multirow{2}{*}{$d_{ff}$} & \multirow{2}{*}{$d_{res}$} & Params.  & \multicolumn{3}{c|}{Perplexity} \\ \cline{7-9}
           Epoch & erged &     &    &     & in M   & Train & Dev & Test\\ \hline \hline
           5.5 & \multirow{4}{*}{Yes} & 12   &  4096   & \multirow{4}{*}{512}   & 268  & 57.3  & 59.9 & 62.3 \\ \cline{1-1} \cline{3-4}
           \multirow{3}{*}{5} & & 24   &  \multirow{3}{*}{2048} & & 281     & 55.6   & 58.0 & 60.7 \\ 
           & & 32  &   & &   306  &   53.4 & 56.6 &  59.5 \\ 
           & & 42  &   & &   338   &   51.2 & \textbf{55.0} &  \textbf{57.7} \\ \hline \hline
          \multirow{2}{*}{3} & \multirow{2}{*}{No} &  80 & \multirow{2}{*}{2048}   & \multirow{2}{*}{512}   & 380& 51.9   & 54.3  & 56.9   \\ 
            &   &  96 &    &    & 431   & 50.9  & \textbf{53.7} & \textbf{56.3}  \\ \hline
       \end{tabular}
%		\begin{tabular}{|c|c|c|c|c||r|r|r|} \hline
%			\multirow{2}{*}{$L$}  &  \multirow{2}{*}{$d_{ff}$} & Para. & Max  & Conv-& \multicolumn{3}{c|}{Perplexity} \\ \cline{6-8}
%			&    &    in M & Ep. & erged  & Train & Dev & Test\\ \hline
%			12   &  4096      &268 & 55 & \multirow{2}{*}{Yes} & 57.3  & 59.9 & 62.3 \\ \cline{1-2}
%			24   &  \multirow{2}{*}{2048}  & 281 & 50  &   & 55.6   & \textbf{58.0} & \textbf{60.7} \\ \cline{4-8} \cline{4-8}
%			32	&   &   306 & 38 & No  & 55.1 & 57.3 & 60.2\\ \hline
%		\end{tabular}
%	}
	\end{center}
	\vspace{-2mm}
\end{table}
\subsection{Parameter Tying}
%   	\vspace{-1mm}
Dehghani et al.~\cite{dehghani2018universal} reports Universal Transformers to perform particularly well
for language modeling. This motivates us to experiment with parameter sharing across layers.
For such models to have comparable number of parameters with the standard deep Transformers,
the dimensions in each layer must be increased, which results in slower training; here we simply investigate the effect
of number of recurrence.
Table \ref{tab:univ} shows the perplexity results.
%This is interesting since we can potentially get improvements without additional parameters. 
First of all, we observe that the model performance is behind that of the standard Transformer\footnote{We note that here the direct comparison is not as straightforward as between the standard Transformers. In fact, we observe that the training hyperparameters tuned
for the standard Transformers can not be directly applied to Universal Transformers; specifically, we find it crucial to reduce the gradient norm clipping threshold from 1 to 0.1, which is potentially slowing down the convergence.} (Table \ref{tab:depth_ep25}).
However, we clearly observe that increasing the number of layers from 3 to 12 consistently improves the perplexity.
This improvement without additional parameters motivates future work to investigate further parameter sharing strategies for Transformers.
    	\vspace{-2mm}
        	    \begin{table}[!ht]
        	    	\caption{\label{tab:univ} {\it Perplexity \textbf{after 2.5 epoch} for ($L$, 8192, 1024, 16) models with shared parameters across all layers.}}
        	    	        	    	\vspace{-1mm}
        	    	%    	\vspace{-7mm}
        	    	\begin{center}
        	    		%\resizebox{\columnwidth}{!}{
        	    		\begin{tabular}{|c|c||r|r|} \hline
        	    			 \multirow{2}{*}{$L$} &Params. & \multicolumn{2}{c|}{Perplexity}  \\ \cline{3-4}
        	    			             & in M & Train & Dev \\ \hline
        	    			  3    & \multirow{3}{*}{329}  & 82.6  &79.9 \\ 
        	    			  6   &                                              & 76.7 & 74.6 \\  
        	    			  12   &                                             & \textbf{74.2} &  \textbf{72.1} \\  \hline
        	    		\end{tabular}
        	    	\end{center}
        	    	% univ_12.re_transfo_d00.8192_1024.sgd.cl01.lr1.16_heads /work/asr3/irie/experiments/lm/librispeech/018-02-23--lmword/data-train/univ_3.re_transfo_d00.8192_1024.sgd.cl01.lr1.16_heads 13
        	    	% /work/asr3/irie/experiments/lm/librispeech/018-02-23--lmword/data-train/univ_6.re_transfo_d00.8192_1024.sgd.cl01.lr1.16_heads 14
        	    	\vspace{-2mm}
        	    \end{table}
\begin{figure*}[ht]
	\subfloat[First layer with PE]{
		\centering
		\includegraphics[width=.19\linewidth]{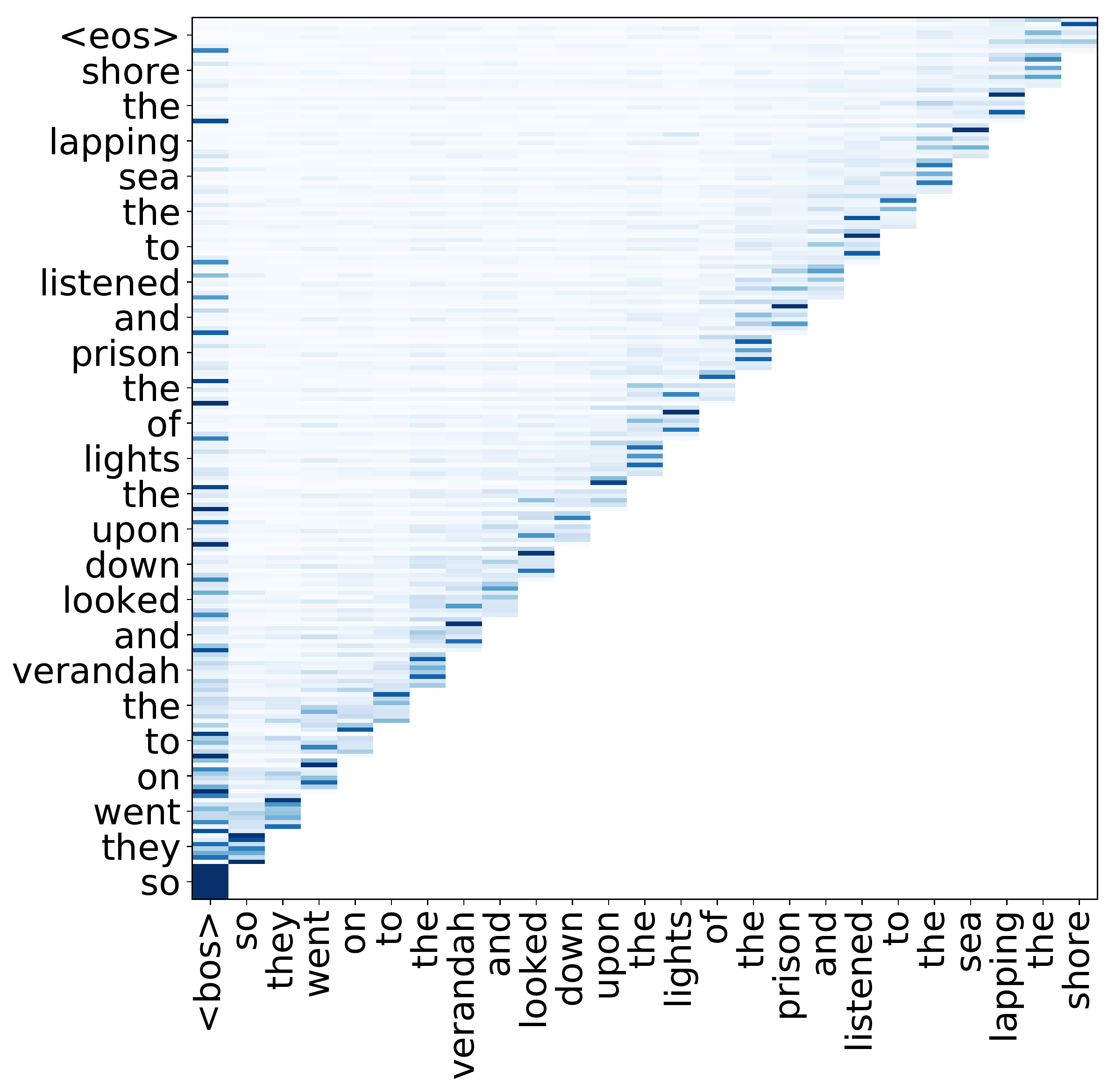}
		\label{sfig:ex1}
	}
	\subfloat[First layer without PE] {
		\centering
		\includegraphics[width=.19\linewidth]{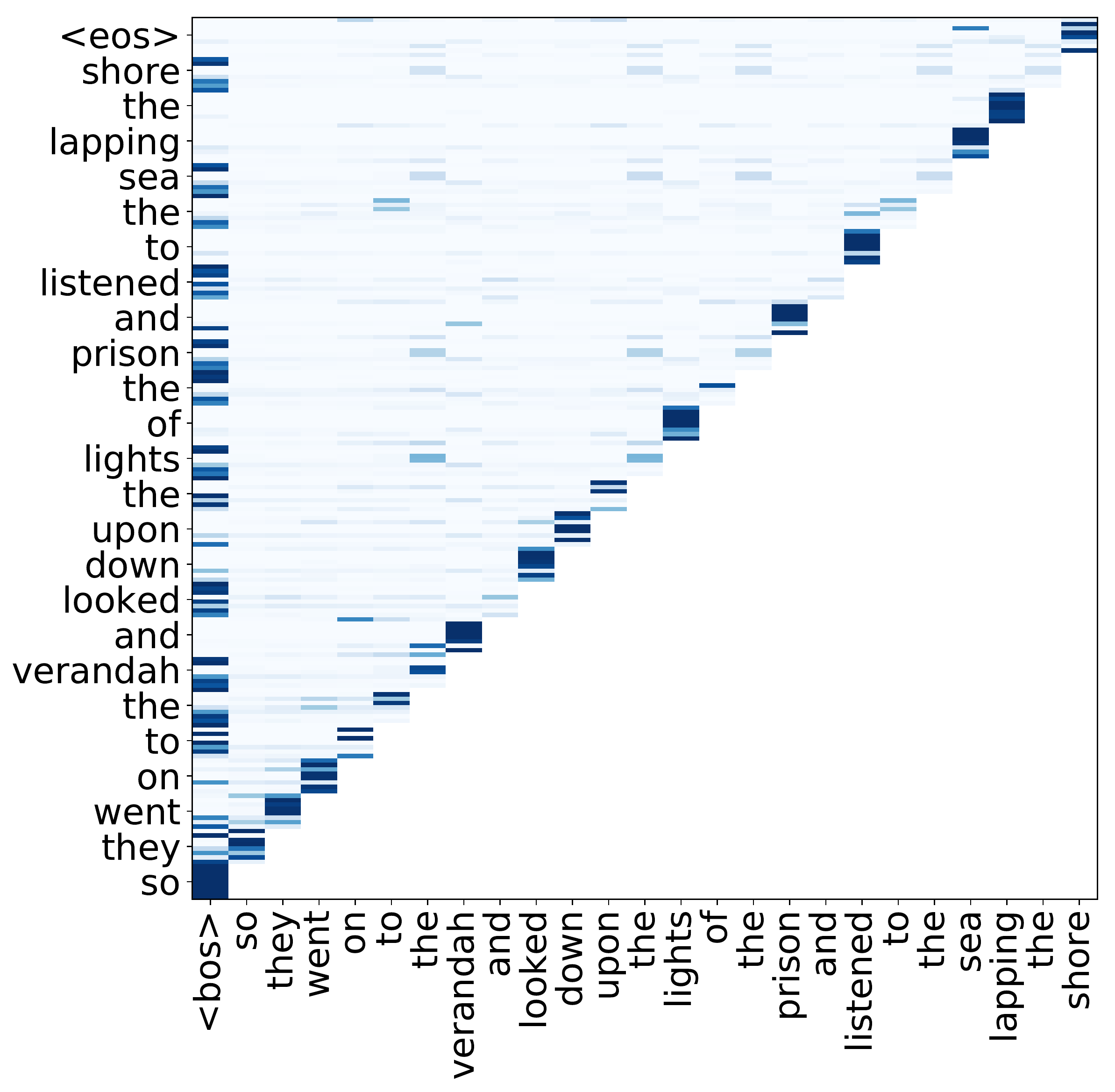}
		\label{sfig:ex2}
	}
	\subfloat["Blur" layer]{
		\centering
		\includegraphics[width=.19\linewidth]{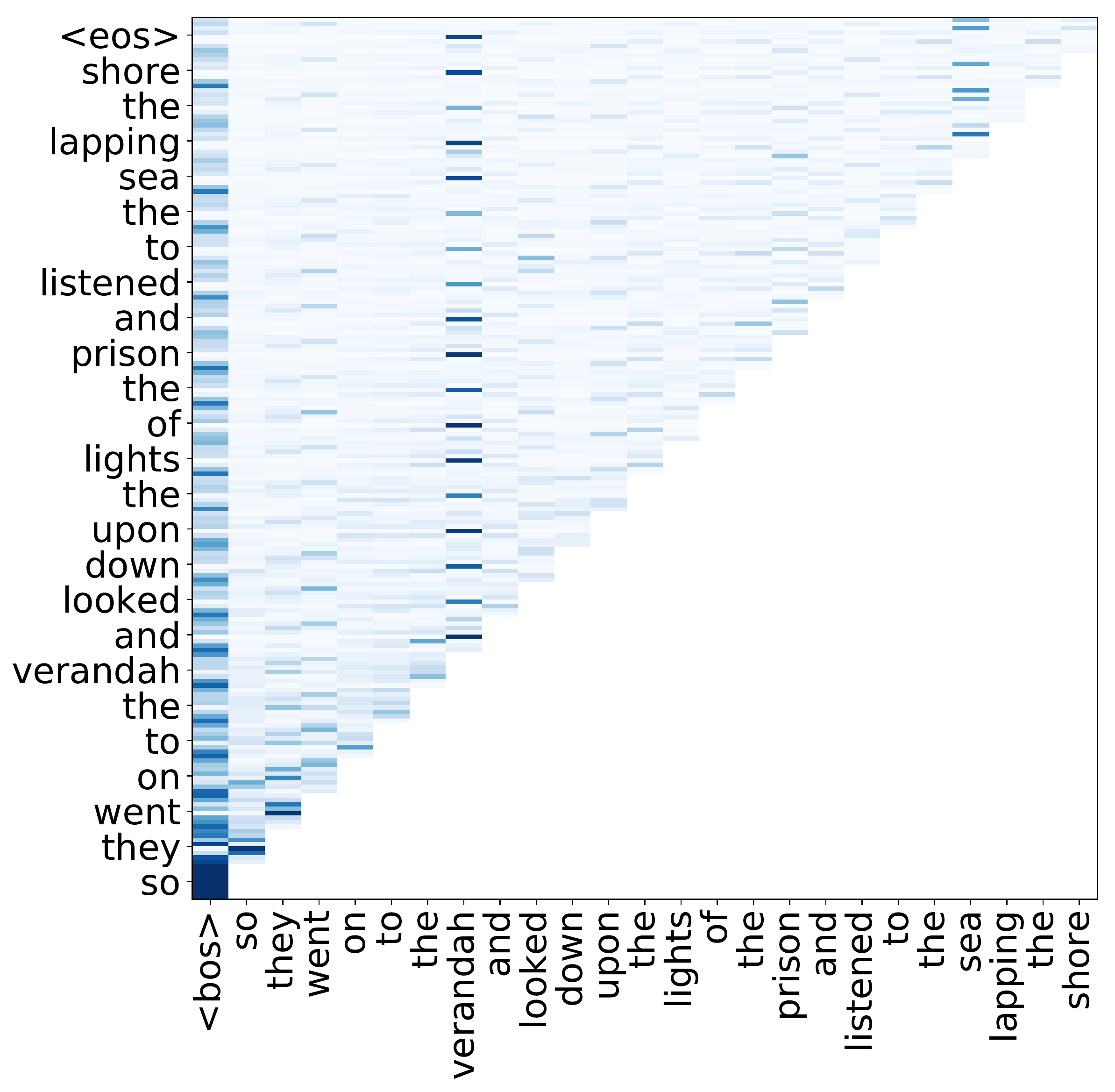}
		\label{sfig:ex1}
	}
	\subfloat["Window" layer] {
		\centering
		\includegraphics[width=.19\linewidth]{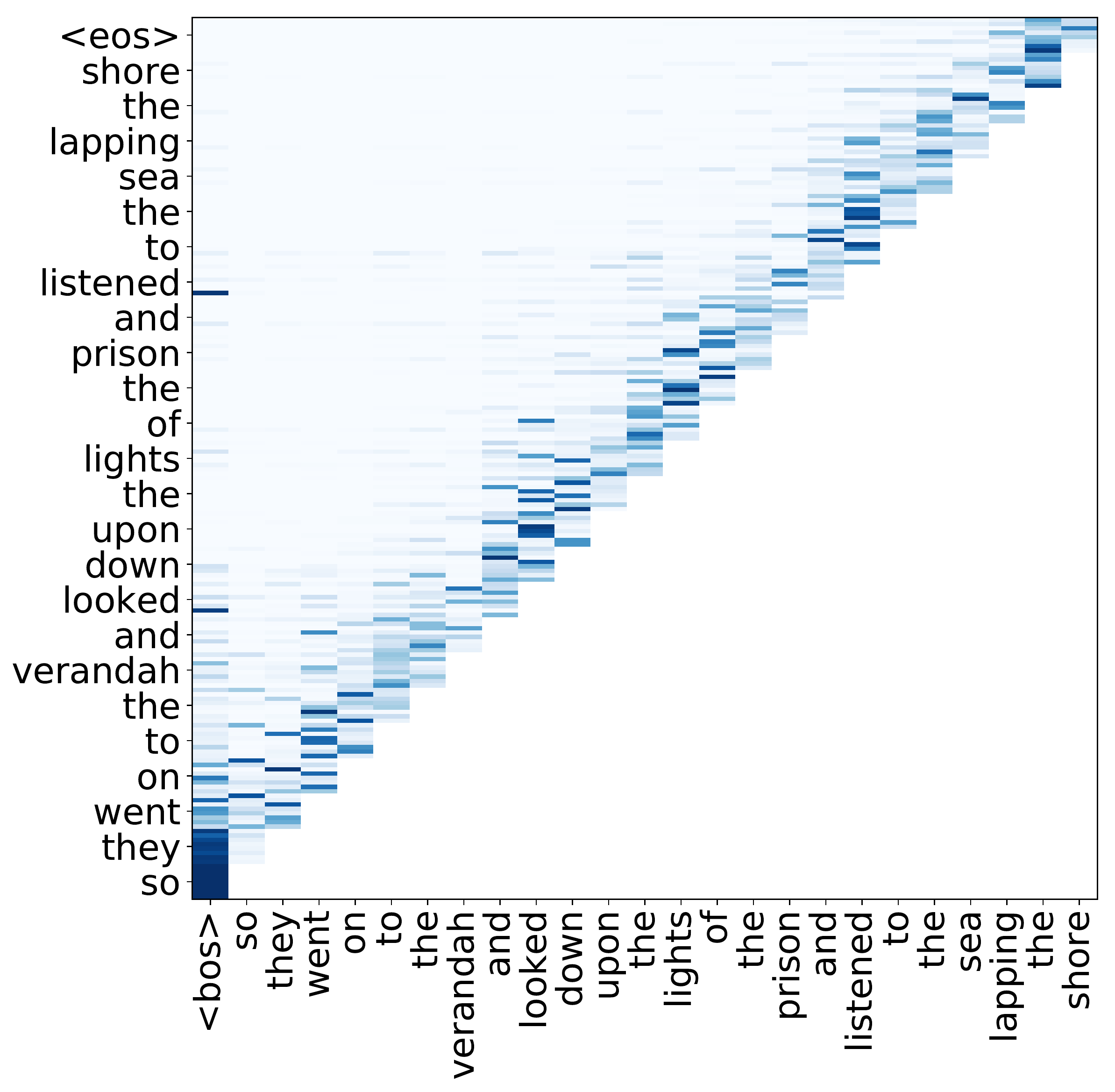}
		\label{sfig:ex2}
	}
	\subfloat["Structured" layer] {
		\centering
		\includegraphics[width=.19\linewidth]{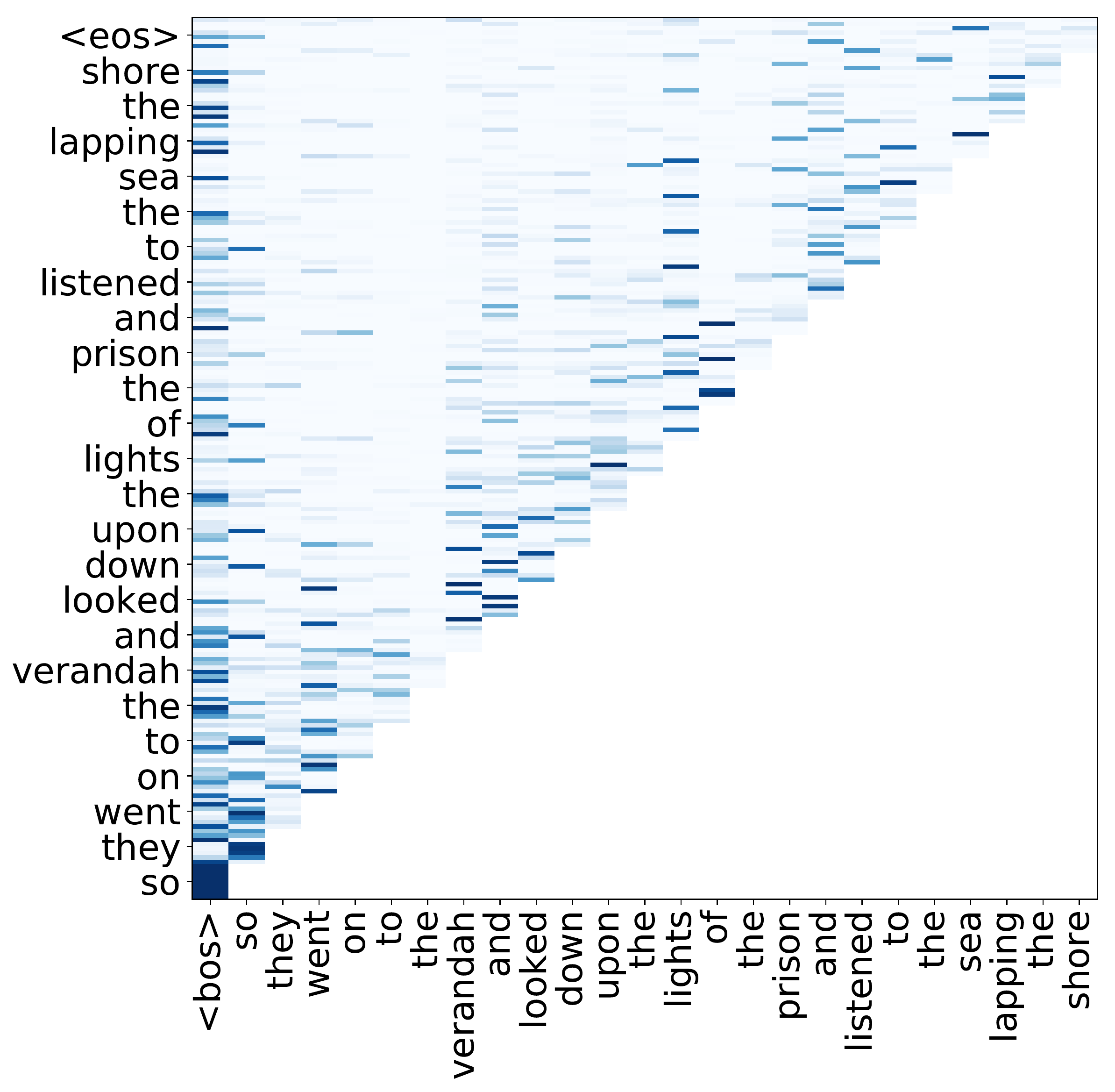}
		\label{sfig:ex2}
	}
	%\vspace{-3mm}
	\caption{\it Layer categories in word-level 24-layer Transformer language models.
		The x-axis corresponds to the input words. The y-axis shows the target words;
		each target word position has 8 sub-rows corresponding to 8 heads.
		"PE" denotes positional encoding.}
	\label{fig:heatmaps}
		\vspace{-6mm}
\end{figure*}
%The hyper-parameters had to be reconsidered/tuned. Reduce clipping value.
\vspace{-2mm}
\section{ASR Experiments}
%\vspace{-1mm}
\subsection{Lattice Rescoring Results}
%\vspace{-1mm}
%\vspace{-2mm}
We apply our word-level Transformer language models to conventional hybrid speech recognition by lattice rescoring.
The standard push-forward lattice rescoring algorithm \cite{sundermeyer2014:rescoring} for long-span language models
can be directly applied to self-attention based models. The only modifications from the RNN version
is to define the "state" as all hidden states ($h_t^{(l)}$ in Sec.\ref{sec:model}) in all layers from all predecessor positions and the current position ($t$; for position encoding). Table \ref{hybrid} shows the WERs and perplexities (PPL). Our baseline acoustic model is based on multi-layer
bi-directional LSTM \cite{zeyer17:lstm}. Further descriptions of our baseline acoustic model can be found in \cite{luescher19}.
We obtain consistent improvements in terms of WER over the LSTM baselines.
\vspace{-1mm}
        \begin{table}[h]
        	\setlength{\tabcolsep}{0.1em}
        	\centering
        	\caption{\it WERs (\%) for \textbf{hybrid} systems on the LibriSpeech \textbf{960hr}. 4-gram model is used in
        		the first pass to generates lattices for rescoring. The row "Lattice" shows oracle WERs of the lattices.}
        	%	\vspace{-4mm}
        	\label{hybrid}
        	%       \setlength{\tabcolsep}{0.35em}
        	%       \centering
        	\begin{tabular}{ |l|c|c||r|r|r|r|r|r|r|r|} \hline
        		\multirow{3}{*}{LM} & \multirow{3}{*}{$L$} & \multirow{2}{*}{Para.} & \multicolumn{4}{|c|}{dev} & \multicolumn{4}{|c|}{test}    \\ \cline{4-11}
        		&  & \multirow{2}{*}{in M} &  \multicolumn{2}{|c|}{clean} & \multicolumn{2}{|c|}{other} & \multicolumn{2}{|c|}{clean} & \multicolumn{2}{|c|}{other}    \\ \cline{4-11}
        		& & & PPL & WER & PPL & WER & PPL & WER & PPL & WER \\ \hline
        		4-gram    & - & 230 & 151.7 & 3.4 & 140.6 & 8.3 & 158.1 & 3.8 & 145.7 & 8.8 \\ 
        		Lattice   & - & - & - & 1.0 & - & 2.3 & - & 1.3 & - & 2.6  \\ \hline
        		LSTM &  2 & 1048 & 60.2 & 2.3 & 60.2 & 5.4 & 64.8 & 2.6 & 61.7 & 5.9  \\ \hline
        		\multirow{2}{*}{Trans-}  & 24  & 281 & 57.8 & 2.2 & 58.3  & \textbf{5.2}  & 62.2 & \textbf{2.5} & 59.4 &  5.7  \\ 
        		\multirow{2}{*}{former} & 42 & 338  & 54.5 & \textbf{2.1} & 55.5 & \textbf{5.2} & 59.1 & \textbf{2.5} & 56.4 & 5.7  \\ 
        		& 96 & 431  & \textbf{53.2} & \textbf{2.1} & \textbf{54.2} & \textbf{5.2} & \textbf{57.6} & \textbf{2.5} & \textbf{55.0} & \textbf{5.6} \\ \hline
        	\end{tabular}
        	\vspace{-1mm}
        \end{table}
\subsection{End-to-End ASR Shallow Fusion Results}
We train 10K BPE-level Transformer language models to be combined with an attention-based encoder-decoder speech model by shallow fusion \cite{gulcehre+al-2016-monolingual, toshniwal2018comparison}.
The 10K BPE level training data has a longer average length of 24 tokens per sentence with the longest sentence length
of 1343, which is still manageable without any truncation for self-attention.
We use the Transformer architecture of (24, 4096, 1024, 8).
The LSTM model has 4 layers with 2048 nodes.
We refer to our previous work \cite{zeyer2018:asr-attention} for the description of the baseline attention model; the baseline WERs better than our previous work \cite{zeyer2018:asr-attention} are obtained by improved curriculum learning and longer training.
%\cite{zeyer2018:attanalysis, zeyer2018:asr-attention, doetsch2017:returnn, zeyer2018:returnn}
Table \ref{e2e} shows both perplexities and WERs. Following \cite{hannun2019sequence},
	we introduce an end-of-sentence penalty in shallow fusion to benefit from a large beam size of 64.
Again, we obtain consistent improvements over the LSTM baseline.
These results are better than previously reported WERs \cite{hannun2019sequence, zeghidour2018fully, irie2019}
for end-to-end models without data augmentation \cite{park2019specaugment}.
\begin{table}[h]
	\vspace{-1mm}
	\setlength{\tabcolsep}{0.3em}
	\centering
%	\footnotesize
	%       \vspace{-2mm}
	\caption{\it WERs (\%) for \textbf{attention-based models} on LibriSpeech \textbf{960hr} dataset.
		Perplexities are on the 10K BPE level.}
	%	\vspace{-4mm}
	\label{e2e}
	%	\resizebox{\columnwidth}{!}{
	\begin{tabular}{ |l|l|r|r|r|r|r|r|r|r|} \hline
		\multirow{3}{*}{LM} &  \parbox[t]{3mm}{\multirow{3}{*}{\rotatebox[origin=c]{90}{Beam}}}& \multicolumn{4}{|c|}{dev} & \multicolumn{4}{|c|}{test}    \\ \cline{3-10}
		&  &  \multicolumn{2}{|c|}{clean} & \multicolumn{2}{|c|}{other} & \multicolumn{2}{|c|}{clean} & \multicolumn{2}{|c|}{other}    \\ \cline{3-10}
		& & PPL & WER & PPL & WER & PPL & WER & PPL & WER \\ \hline
		None & 12 & - & 4.3 & -& 12.9 &- & 4.4 & - & 13.5  \\ \hline
		LSTM & \multirow{2}{*}{64}  &  43.7 &   2.9  & 46.4 &  8.9 & 47.1  & 3.2  & 47.2 &  9.9   \\ 
		Transfo. & & \textbf{35.9} & \textbf{2.6} & \textbf{38.9} & \textbf{8.4} & \textbf{38.8} & \textbf{2.8} & \textbf{39.0}  &  \textbf{9.3}  \\ \hline
	\end{tabular}
%	\begin{tabular}{ |l|l|r|r|r|r|r|r|r|r|} \hline
%		\multirow{3}{*}{LM} &  \parbox[t]{3mm}{\multirow{3}{*}{\rotatebox[origin=c]{90}{Beam}}}& \multicolumn{4}{|c|}{dev} & \multicolumn{4}{|c|}{test}    \\ \cline{3-10}
%		&  &  \multicolumn{2}{|c|}{clean} & \multicolumn{2}{|c|}{other} & \multicolumn{2}{|c|}{clean} & \multicolumn{2}{|c|}{other}    \\ \cline{3-10}
%		& & PPL & WER & PPL & WER & PPL & WER & PPL & WER \\ \hline
%		None & 12 & - & 4.3 & -& 12.9 &- & 4.4 & - & 13.5  \\ \hline
%		LSTM & \multirow{2}{*}{64}  & 47.9  &  3.3    & 50.7 & 9.7 & 51.1 & 3.5  & 51.1 & 10.7    \\ 
%		Transfo. & & \textbf{40.2} & \textbf{3.0} & \textbf{43.2} & \textbf{9.4} & \textbf{43.0} & \textbf{3.3} & \textbf{43.5} & \textbf{10.3} \\ \hline
%	\end{tabular}
	% ADD PATHS HERE
	% We select model checkpoints which give the best perplexity in the Dev set.
	% The shallow fusion weights are optimized on dev-clean and dev-other separately and used for decoding test-clean and test-other respectively.
	% beam 12:
	% /work/asr3/irie/experiments/lm/librispeech/2018-03-12-zeyer-2018-02-26--att-lm/data-recog/base3.retrain2.transfo_lm_ep23.166/scoring.wers
	% /work/asr3/irie/experiments/lm/librispeech/2018-03-12-zeyer-2018-02-26--att-lm/data-recog/base3.retrain2.lstm_lm.166/scoring.wers
	%	}
	%	\vspace{-4mm}
\end{table}
\vspace{-2mm}
\section{Analysis}
Compared with hidden states in RNNs, attention weights are easier to be visualized, which gives opportunity
for analysis. In particular, we focus on the comparison of the Transformer language models
with and without positional encoding.
%The plots are generated using our best models with configuration (24, 2048, 512, 8).
%We also illustrate that, as opposed to attentions in encoder-decoder modeling, attention in language modeling
%is more about feature engineering and detection, rather than the exact \textit{alignment}.
%In another words, it's rather a (deep) bag-of-words with adaptive weights than
%the word trigger.
\subsection{Transformer LM without positional encoding}
In the autoregressive problem where a new token is provided to the model at each time step,
the amount of information the model has access to strictly increases from left to right at the lowest level of the network;
the deeper layers should be able to recognize this structure which should provide the model with some positional information by its own.
To check this hypothesis, we train models without any positional encoding.
First, we observe that they give better perplexities than the models with sinusoidal positional encoding (Table \ref{tab:nopos}).
    \begin{table}[!ht]
	%\vspace{-5mm}
%			\setlength{\tabcolsep}{0.4em}
\caption{\label{tab:nopos} {\it Effect of sinusoidal positional encoding. Perplexity \textbf{after 5 epochs} (13M updates) for ($L$, 2048, 512, 8) models.}}
	    	\vspace{-1mm}
	\begin{center}
		%\resizebox{\columnwidth}{!}{
		\begin{tabular}{|c|c|c||r|r|r|} \hline
			\multirow{2}{*}{$L$} & Position. & Params. & \multicolumn{3}{c|}{Perplexity} \\ \cline{4-6}
			& encoding & in M. & Train & Dev & Test \\ \hline
			\multirow{2}{*}{12}  & Sinusoidal  & \multirow{2}{*}{243} & 61.8  & 63.1 & 66.1 \\
			& None             & & 58.0 & \textbf{60.5}  & \textbf{63.4}  \\ \hline
			\multirow{2}{*}{24}  & Sinusoidal & \multirow{2}{*}{281}  & 55.6 & 58.0 & 60.8  \\ 
			& None & & 52.7 & \textbf{56.6} & \textbf{59.2}  \\ \hline
            \multirow{2}{*}{42}  & Sinusoidal & \multirow{2}{*}{338}  & 51.2 & 55.0 &  57.7   \\
            & None & & 50.5 & \textbf{54.2} & \textbf{56.8}  \\ \hline
		\end{tabular}
	\end{center}
	% /work/asr3/irie/experiments/lm/librispeech/018-02-23--lmword/data-train/univ_3.re_transfo_d00.8192_1024.sgd.cl01.lr1.16_heads 13
	% /work/asr3/irie/experiments/lm/librispeech/018-02-23--lmword/data-train/univ_6.re_transfo_d00.8192_1024.sgd.cl01.lr1.16_heads 14
	    	\vspace{-2mm}
\end{table}
%\vspace{-5mm}

\subsection{First layer}
%	\vspace{-1mm}
\label{sec:first_layer}
The attention in the first layer is the most straightforward for interpretation because
the feature at each position exactly corresponds to the word at the position (while deeper layers can potentially
shuffle the feature content).
The attention weights in the first layer of 24-layer Transformer language models with and without positional encodings
are visualized in Figure \ref{fig:heatmaps}.
We observe that the first layer of the model with positional encoding (Figure \ref{fig:heatmaps}(a)) learns to create \textit{n-gram features} (roughly 2 or 3-gram), which indicates that the positional information is directly used.
In contrast, the first layer of the model without positional encoding learns to focus on the new input token as can be seen
as the diagonal in Figure \ref{fig:heatmaps}(b) (interestingly, we also see that it ignores some functional words such as "the", "and", "to" which might be modeled by some off-set values, therefore attending to the beginning of sentence token instead),
which demonstrates that the model is aware of the position of the new input.
%We also observe that in each of these layers, many heads are not active (attending to the beginning of sentence token).
%Template solution for language modeling.

%\begin{figure}[b]
%	\hspace{-5mm}
%	\subfloat["Blur" layer]{
%		\centering
%		\includegraphics[width=.35\linewidth]{figures/finals/without/nocaption_att_lay_1.pdf}
%		\label{sfig:ex1}
%	}
%	\subfloat["Window" layer] {
%		\centering
%		\includegraphics[width=.35\linewidth]{figures/finals/without/nocaption_att_lay_4.pdf}
%		\label{sfig:ex2}
%	}
%	\subfloat["Structured" layer] {
%		\centering
%		\includegraphics[width=.35\linewidth]{figures/finals/without/nocaption_att_lay_23.pdf}
%		\label{sfig:ex2}
%	}
%	\vspace{-3mm}
%	\caption{\it Attention weights in the \textbf{first layer}.
%		The x-axis corresponds to the input word. The y-axis shows the target words; each target word position, 8 heads are shown.}
%	\label{fig:heatmaps}
%	\vspace{-5mm}
%\end{figure}
\subsection{Other layers}
%	\vspace{-1mm}
We observe that the behavior of other layers are rather similar for both Transformer models with and without positional encoding.
We find 3 categories of layers in the other 23 layers; the second and third layers are \textit{"blur"} layers as shown in Figure \ref{fig:heatmaps}(c), which seems to roughly average
over all positions (while we can also see that some heads focus on \textit{difficult} words, here "verandah").
Layer 4 to 9 are \textit{window} layers which focus on the local n-gram. A representative example is show in Figure \ref{fig:heatmaps}(d). 
Finally, we find the top layers 10 to 24 to be more \textit{structured}, attending to some specific patterns; an example is shown in Figure \ref{fig:heatmaps}(e).

\vspace{-0.5mm}
\section{Conclusion}
\vspace{-1mm}
We apply deep Transformer language models for speech recognition.
We show that such models outperform the shallow stack of LSTM-RNNs on both word-level and BPE-level modeling.
Future work investigates application of crucial components of deep Transformers (such as layer normalization)
to deeper LSTM models; e.g., the RNMT+ \textit{decoder} architecture \cite{bestof18} for
language modeling. Furthermore, we do not apply any regularization on models for the LibriSpeech task,
as no overfitting is observed in the range of model sizes we experimented with (for the word-level models).
We can possibly still improve our models simply by scaling up their size and using regularization.

\vspace{1.5mm}
\setlength\parindent{65pt}
%\section{Acknowledgements
{\large \textbf{Acknowledgements}} \vspace{1mm} \\
\setlength{\intextsep}{2pt}%
\setlength{\columnsep}{2pt}%
\footnotesize
%\scriptsize
%\small
%\tiny
%\begin{wrapfigure}{r}{0.15\textwidth}
%\includegraphics[width=0.15\textwidth]{figures/euerc.png}
%\end{wrapfigure}
This work has received funding from the European Research Council (ERC) under the European Union's Horizon 2020 research and innovation programme (grant agreement No 694537, project "SEQCLAS") and from a Google Focused Award. The work reflects only the authors' views and none of the funding parties is responsible for any use that may be made of the information it contains.
We thanks Liuhui Deng for contributing to our lattice rescoring code, Arne Nix and Julian Schamper for sharing their base Transformer configs, as well as Eugen Beck, Christoph L\"uscher and Wei Zhou for help with generating lattices. Experiments were partially 	performed with computing resources
granted by RWTH Aachen University under project nova0003.
\clearpage
\newpage

\bibliographystyle{IEEEtran}
\def\thebibliography#1{\begin{center}
		{\bf \large References \vspace{-.5em}\vspace{2pt}}
	\end{center} \eightpt \list
	{[\arabic{enumi}]}{\settowidth\labelwidth{[#1]}\leftmargin\labelwidth
		\advance\leftmargin\labelsep
		\usecounter{enumi}}
%	\def\newblock{\hskip .11em plus .33em minus .07em}
%	\sloppy\clubpenalty4000\widowpenalty4000
%	\sfcode`\.=1000\relax
}
\let\endthebibliography=\endlist

\let\normalsize\footnotesize\normalsize
\SetTracking{encoding=*}{-15}\lsstyle  % still somewhat ok
%\SetTracking{encoding=*}{-85}\lsstyle

%\renewcommand{\baselinestretch}{0.1}\normalsize
% http://tex.stackexchange.com/questions/93859/condense-the-space-between-bibliographic-entries
\let\OLDthebibliography\thebibliography
\renewcommand\thebibliography[1]{
	\OLDthebibliography{#1}
	\setlength{\parskip}{2pt}
	\setlength{\itemsep}{1pt plus 0.07ex}
}
% \eightpt
% \bibliographystyle{IEEEtran}
% \renewcommand{\baselinestretch}{0.7}\normalsize
  \bibliography{paper}
%\bibliographystyle{IEEEtran}

%% Give us some more space:
%% Tune this later if needed, or just uncomment if not needed.
%
%%\def\baselinestretch{0.8}
%\let\normalsize\footnotesize\normalsize
%\SetTracking{encoding=*}{-15}\lsstyle  % still somewhat ok
%%\SetTracking{encoding=*}{-85}\lsstyle
%
%\renewcommand{\baselinestretch}{0.1}\normalsize
% http://tex.stackexchange.com/questions/93859/condense-the-space-between-bibliographic-entries
%\let\OLDthebibliography\thebibliography
%\renewcommand\thebibliography[1]{
%	\OLDthebibliography{#1}
%	\setlength{\parskip}{0pt}
%	\setlength{\itemsep}{0pt plus 0.07ex}
%}

%\bibliography{mybib}

\end{document}